\documentclass{article}

\usepackage{microtype}
\usepackage{graphicx}
\usepackage{subcaption}
\usepackage{booktabs} 
\usepackage{float}

\usepackage{tikz}
\usetikzlibrary{arrows.meta, positioning, shapes, calc, decorations.pathreplacing}
\usepackage{pgfplots}
\pgfplotsset{compat=1.18}
\usepgfplotslibrary{fillbetween} 
\usepackage{multirow}

\usepackage{inconsolata} 
\usepackage{hyperref}

\usepackage[preprint]{icml2026}

\usepackage{amsmath}
\usepackage{amssymb}
\usepackage{mathtools}
\usepackage{amsthm}

\usepackage[capitalize,noabbrev]{cleveref}

\theoremstyle{plain}

\theoremstyle{definition}

\theoremstyle{remark}

\usepackage[textsize=tiny]{todonotes}

\begin{document}

\twocolumn[
  \icmltitle{Hidden Heroes and Gradient Bloats: Layer-Wise Redundancy Inverts Attribution in Transformers}

  \icmlsetsymbol{equal}{*}

  \begin{icmlauthorlist}
    \icmlauthor{Donald Ye}{yyy}
  \end{icmlauthorlist}

  \icmlaffiliation{yyy}{Department of Computer Science, 
Fu Foundation School of Engineering and Applied Science, 
Columbia University, New York, NY, USA}

  \icmlcorrespondingauthor{Donald Ye}{dy2524@columbia.edu}

  \icmlkeywords{Mechanistic Interpretability, Gradient Attribution, Causal Abstractions}

  \vskip 0.3in
]

\printAffiliationsAndNotice{} 

\begin{abstract}
Gradient-based attribution is the workhorse of 
mechanistic interpretability, yet whether it 
reliably tracks causal importance at the component 
level remains largely untested. We causally 
evaluate this assumption across two algorithmic 
tasks and up to 10 random seeds, uncovering a 
systematic, layer-wise failure: gradient 
attribution consistently overvalues early-layer 
\textbf{Gradient Bloats} and undervalues 
late-layer \textbf{Hidden Heroes}. Rank 
correlation collapses from $\rho = 0.72$ on 
sequence reversal to $0.27$ on sequence sorting, reaching 
$\rho = -0.18$ in individual seeds. This failure stems 
from first-order gradient attribution's 
inability to detect collective redundancy: joint 
Bloat ablation causes $14\times$ greater damage 
than individual results predict. Consequently, 
Bloats dominate gradient rankings despite 
negligible functional impact, while ablating 
Hidden Heroes destroys OOD accuracy 
($-36.4\% \pm 22.8\%$). This systematic 
inversion of early-layer feature extraction and 
late-layer computation motivates causal 
validation as a prerequisite for circuit-level 
claims. Code available at \url{https://github.com/donald-ye/casual-gradient}
\end{abstract}


\section{Introduction}

We test whether gradient magnitude reliably identifies 
the transformer components that causally drive 
out-of-distribution (OOD) generalization. It does 
not---and the failure is structured and predictable.

Gradient-based attribution underpins pruning, circuit 
discovery, and interpretability audits in transformers 
\citep{michel2019sixteen, han2015learning, 
conmy2023automated, sun2023wanda, wang2022interpretability}, 
with large-gradient components treated as functionally 
important. While input-level sanity checks show 
saliency maps can fail to reflect model behavior 
\citep{adebayo2018sanity}, whether gradient magnitude 
correctly identifies which architectural components 
drive actual computation remains largely unvalidated 
against causal interventions. If this assumption fails systematically, pipelines 
that rely on it risk destroying OOD generalization 
by pruning critical components, and circuits 
identified downstream may be artifacts of the 
attribution method rather than genuine computational 
structure \citep{conmy2023automated}. 

The gap is task-dependent and severe. Gradient tracks 
causal importance on sequence reversal ($\rho = 0.72$) 
but collapses on sorting ($\rho = 0.27$), with 
individual seeds reaching $\rho = -0.18$. The failure 
is not random noise: low-gradient \textbf{Hidden 
Heroes} concentrate in later layers while high-gradient 
\textbf{Gradient Bloats} concentrate in earlier 
layers, stable across seeds, baselines, and thresholds. 
The mechanism is redundant circuit structure: joint 
Bloat ablation causes $14\times$ greater damage than 
individual ablation predicts, revealing a compensating 
circuit that first-order gradient attribution 
fundamentally cannot resolve.

\noindent\textbf{Contributions:} 
(\textbf{i})~causal measurement of 
the gradient-causal gap across up to 10 seeds, 2 
tasks, and 2 ablation baselines; (\textbf{ii})~a 
stable layer-wise structure where Hidden Heroes 
concentrate in late layers and Gradient Bloats in 
early layers, robust to baseline and threshold; 
(\textbf{iii})~component identity stability, with 
specific heads occupying consistent roles across 
seeds (L3\_H3 Hero in 7/10; L1\_H1, L1\_H3 Bloats 
in 6/10); and (\textbf{iv})~$14\times$ 
superadditivity in joint Bloat ablation, providing 
mechanistic evidence that redundant circuits drive 
gradient overvaluation.

\begin{figure*}[t]
    \centering
    \includegraphics[width=.95\textwidth]{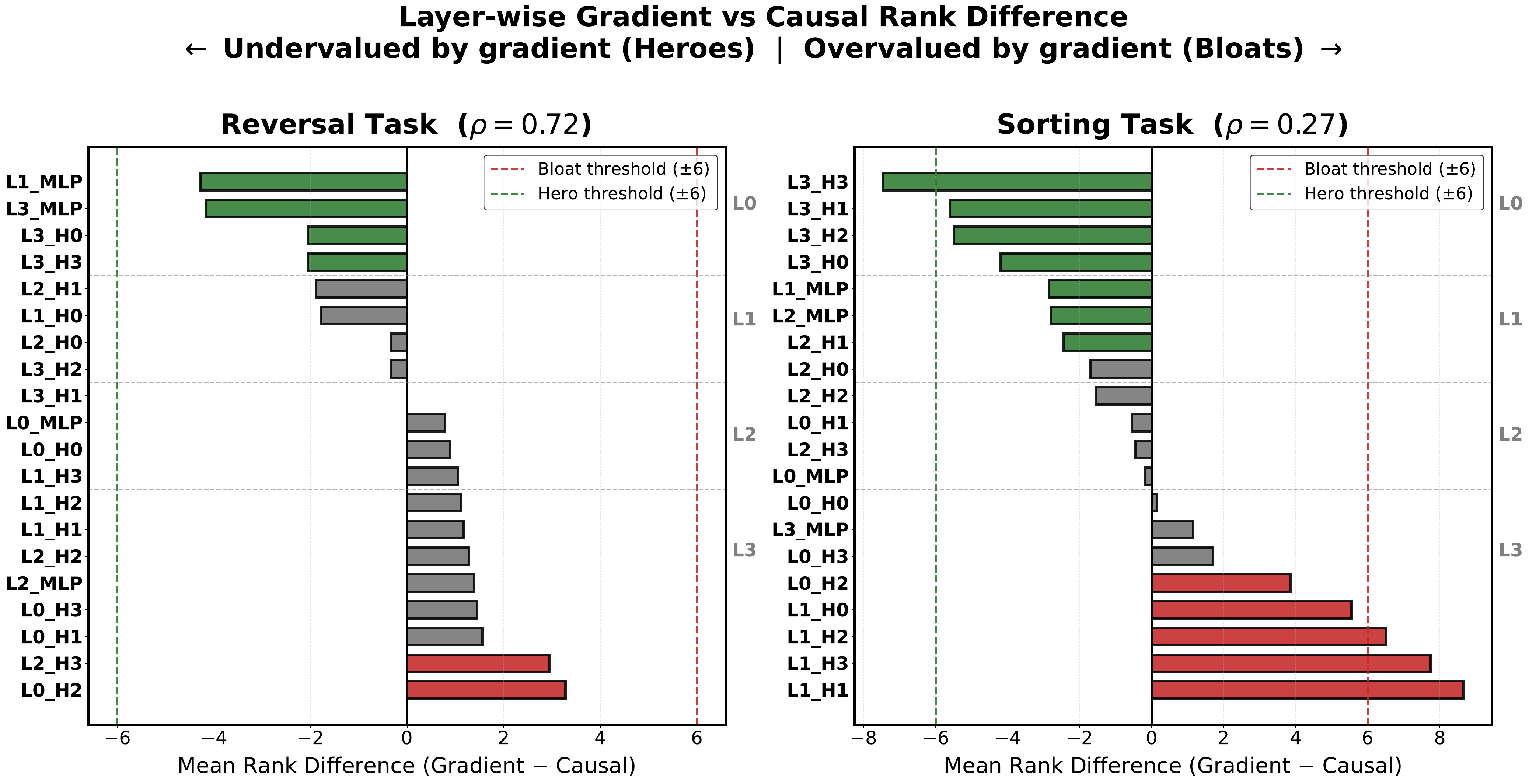}
    \caption{Mean rank difference (Gradient $-$ Causal) 
per component across seeds. Dashed lines mark the 
$\pm 6$ Hero/Bloat thresholds. On sorting (right), 
Layer~1 heads cluster as Bloats and Layer~3 heads 
cluster as Heroes. On reversal (left), the same 
layer-wise pattern exists but is attenuated, 
consistent with the stronger gradient-causal 
alignment ($\rho = 0.72$ vs.\ $0.27$).}
    \label{fig:waterfall}
\end{figure*}

\section{Method}

\subsection{Task and Model Architecture}
We evaluate on two benchmarks: \textit{Sequence 
Reversal} and \textit{Sequence Sorting}, using 
integers $x \in \{1, \dots, 99\}$. We use a 
decoder-only Transformer with $L=4$ layers and $H=4$ 
heads, giving 
$|\mathcal{S}|=20$ components; hyperparameters 
are in Appendix~\ref{sec:appendix}. Algorithmic 
tasks provide a lower bound on attribution failure: 
causal structure is fully measurable and circuits 
are relatively simple, making this the setting 
where gradient attribution should perform best 
\citep{conmy2023automated, varma2023explaining}. 
Failure here suggests the gradient-causal gap will 
be exacerbated in naturalistic settings with greater 
redundancy and complexity.

\subsection{Measuring Importance}
Let $\mathcal{S} = \{H_{\ell,h}\} \cup \{M_\ell\}$, where $H_{\ell,h}$ is the $h$-th head in layer $\ell$ and $M_\ell$ is the MLP in layer $\ell$.

To isolate components for generalization, we evaluate models on OOD sequence lengths $N \in \{8, 9, 10, 11\}$. For each seed, we select the length where the model achieves accuracy within $[20\%, 75\%]$: below $20\%$ the model has not generalized to the test length; above $75\%$ component effects are too small to measure reliably. Accuracy is measured as exact sequence match.

\textbf{Gradient Magnitude ($G$).} For each component $i \in \mathcal{S}$, we compute the normalized average Frobenius norm across 50 OOD batches:
\begin{equation}
    G_i = \frac{1}{B} \sum_{b=1}^{B} \frac{1}{\sqrt{N_i}} \left\| \nabla_{\mathbf{W}_i} \mathcal{L}_b \right\|_F
\end{equation}
where $B=50$, $\mathcal{L}_b$ is the cross-entropy loss for batch $b$, and $N_i$ is the parameter count of component $i$. We normalize by $\sqrt{N_i}$ to ensure fair comparisons across heterogeneous sublayers (e.g., $128 \times 32$ attention matrices versus $128 \times 512$ MLP matrices). Following \citet{elhage2021mathematical}, we isolate $\mathbf{W}_V$ and $\mathbf{W}_{\text{out}}$ to bound our measure to each component's residual stream contribution, preventing confounding from query/key routing dynamics. While gradient magnitude measures sensitivity to parameter perturbations rather than direct output contribution, it remains a dominant pruning heuristic \citep{han2015learning, michel2019sixteen}; we test whether this heuristic holds for OOD generalization.

\textbf{Causal Importance ($C$).} We measure functional necessity via two ablation baselines. \textit{Mean ablation} replaces component $i$'s output with its mean activation $\boldsymbol{\mu}_i$ computed over 50 OOD batches ($C_i^{\text{mean}} = \text{Acc}_{\text{base}} - \text{Acc}_{\text{ablated}(i \to \boldsymbol{\mu}_i)}$). \textit{Zero ablation} replaces the component output with the zero vector ($C_i^{\text{zero}} = \text{Acc}_{\text{base}} - \text{Acc}_{\text{ablated}(i \to \mathbf{0})}$). Both measure the causal effect of removing a component under different counterfactuals. Agreement between baselines rules out results being an artifact of the mean ablation off-state.

\subsection{The Gradient-Causal Gap and Classification}
We quantify misalignment via Spearman correlation $\rho$ between $\mathbf{G}$ and $\mathbf{C}$, and define the Gradient-Causal Gap as the rank difference:
\begin{equation}
    \Delta_i = \text{Rank}(G_i) - \text{Rank}(C_i)
\end{equation}
Components are classified by $\Delta_i$: \textbf{Hidden Heroes} ($\Delta_i \leq -6$) are low-gradient but causally essential; \textbf{Gradient Bloats} ($\Delta_i \geq 6$) are high-gradient with negligible causal impact; Aligned components ($|\Delta_i| < 6$) show agreement between measures. The threshold $|\Delta_i| \geq 6$ represents a rank divergence of at least $30\%$ across our 20-component architecture, isolating the most extreme misalignments. We report sensitivity analysis at $\pm 4$, $\pm 6$, and $\pm 8$ in Appendix~\ref{app:threshold}. To validate these categories causally, we ablate the top-two components per class across up to 10 random seeds and measure the resulting OOD accuracy change.

\section{Results}

\subsection{The Gradient-Causal Gap}

\begin{table}[h]
\centering
\caption{Spearman $\rho$ correlation between gradient magnitude and causal importance. Both baselines show the same collapse on sorting, ruling out mean ablation artifact as an explanation (mean $\pm$ std across 10 seeds).}
\label{tab:rho}
\begin{tabular}{lcc}
\toprule
\textbf{Task} & \textbf{Mean Ablation $\rho$} & \textbf{Zero Ablation $\rho$} \\
\midrule
Reversal & $0.72 \pm 0.08$ & $0.73 \pm 0.07$ \\
Sorting  & $0.27 \pm 0.24$ & $0.34 \pm 0.22$ \\
\bottomrule
\end{tabular}
\end{table}

Gradient attribution tracks causal importance on reversal ($\rho=0.72$) but collapses on sorting ($\rho=0.27$), as shown in Table~\ref{tab:rho}.

The aggregate means, however, understate the severity 
of the failure in individual model runs. In Seed 456, 
we observe $\rho \approx 0.00$ under both baselines, 
indicating that gradient rankings have essentially zero 
predictive power regarding which components are 
functionally necessary. Even more striking is Seed 
2020, which reaches $\rho = -0.18$ under mean 
ablation; here, the gradient signal is \textbf{actively 
inversely predictive} of causal importance. Preliminary checks on activation patching confirm this: $\rho = -0.445$ 
between gradient and patching rankings on the 
best-characterized seed.

\subsection{Layer-wise Organization}

The gap is not uniformly distributed across components; it exhibits a predictable layer-wise structure that persists across seeds, baselines, and thresholds.

\begin{figure*}[t]
    \centering
    \includegraphics[width=0.88\textwidth]{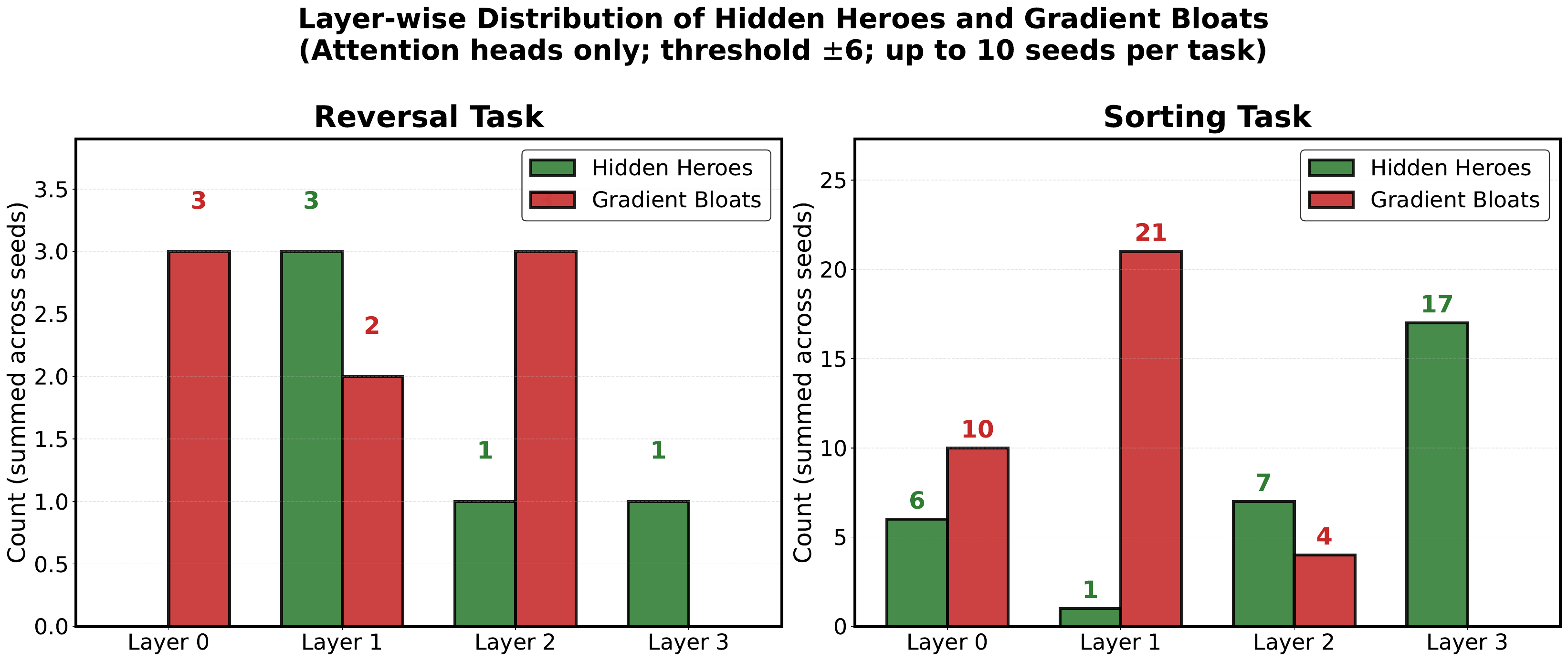}
    \caption{Layer-wise Hero and Bloat counts summed 
    across seeds (attention heads only; threshold 
    $\pm 6$). Layer~1 accumulates 21 Gradient Bloats 
    while Layer~3 accumulates 17 Hidden Heroes on 
    sorting. Pattern is attenuated on reversal, 
    consistent with the weaker gradient-causal gap.}
    \label{fig:layerwise}
\end{figure*}

\paragraph{Layer distribution.} As shown in Figure~\ref{fig:layerwise}, sorting exhibits strong layer dependence: Layer~1 accumulates Gradient Bloats (21 at $\pm 6$), while Layer~3 accumulates Hidden Heroes (17 at $\pm 6$). This pattern is attenuated on reversal ($\rho = 0.72$ vs.\ $0.27$), robust under both baselines, and stable across thresholds $\pm 4$, $\pm 6$, and $\pm 8$ (Appendix~\ref{app:threshold}).

\paragraph{Component identity stability.}
Specific heads occupy stable roles across random initializations. For example, L3\_H3 is classified as a Hidden Hero in 7 out of 10 sorting seeds, while L1\_H1 and L1\_H3 consistently emerge as Gradient Bloats in 6 out of 10 seeds. This suggests real functional circuit roles rather than stochastic noise.

\paragraph{Layer gradient norm distribution.}
Raw gradient norms decrease monotonically across 
layers on sorting (Layer~0: $0.139$, Layer~1: 
$0.172$, Layer~2: $0.093$, Layer~3: $0.052$), 
providing a structural explanation for early-layer 
Bloat concentration: architectural position 
determines gradient magnitude independently of 
causal importance (Appendix~\ref{app:gradnorms}).

\subsection{Pruning Consequences}

To confirm these consequences, we ablate each class 
and measure OOD accuracy change across seeds.

\paragraph{Reversal.}
Ablating Hidden Heroes causes a mean OOD accuracy 
change of $-36.4\% \pm 22.8\%$, devastating and 
directionally consistent across all seeds despite 
high variance. Ablating Gradient Bloats causes only 
$-10.1\% \pm 10.4\%$. Heroes cause $3.6\times$ 
greater damage when removed, yet gradient attribution 
ranks them as less critical.

\paragraph{Sorting.}
Ablating Hidden Heroes causes a change of $-13.9\% \pm 9.4\%$. Ablating Gradient Bloats causes a larger but highly variable change of $-25.4\% \pm 23.2\%$. This seemingly paradoxical result is resolved by the redundant circuit structure described below.

\paragraph{Superadditivity and the Redundancy Paradox.}
The sorting results are explained by a striking 
\textit{superadditivity} effect. When Gradient Bloats 
are ablated \textbf{individually}, the mean accuracy 
change is a negligible $3.1\%$ per component. However, 
when the same Bloats are ablated \textbf{jointly}, 
accuracy collapses by $43.8\%$, representing a ratio 
$14\times$ greater than what individual results predict 
(Appendix~\ref{app:superadditivity}). 

This reveals the mechanism: Bloats form a distributed, redundant circuit where individual components compensate for one another. First-order gradient attribution assigns high signal to all active components, failing to distinguish individually-redundant from individually-critical units. The $\pm 23.2\%$ pruning variance therefore reflects seed-specific circuit topology, not measurement noise.

\section{Discussion}

\subsection{Why the Gap Emerges}

The layer-wise failure reflects a circuit competition 
account: early layers learn broad, high-gradient 
feature extractors, while late 
layers implement sparse, causally indispensable logic 
. The $3.3\times$ gradient differential 
from Layer~1 ($0.172$) to Layer~3 ($0.052$) precisely 
tracks this split. Our claim is that 
backpropagation's inherent structural asymmetry 
systematically misidentifies the drivers within our experimental setting of OOD 
generalization, as confirmed by our pruning and 
superadditivity results. This aligns with grokking 
dynamics, where late-layer generalization circuits 
compete with high-gradient early representations 
\citep{varma2023explaining}. Task complexity dictates 
the severity of this gap: reversal distributes simple 
positional mapping evenly across layers, whereas 
sorting demands cross-token comparisons concentrated 
in late layers, sharply exacerbating the Hero/Bloat 
divide.

The $14\times$ superadditivity finding adds a second 
mechanism: Bloats form redundant compensating circuits 
where each component receives gradient signal even 
though none is individually indispensable. Gradient 
attribution, as a first-order method measuring only 
local loss sensitivity, cannot distinguish collective 
redundancy from genuine importance, producing 
systematic overvaluation of every component in the 
redundant circuit simultaneously. We propose this as 
a hypothesis; direct measurement of circuit formation 
dynamics during training would be required to confirm 
it.

\subsection{Relation to Prior Work}
Prior sanity checks show saliency maps fail to reflect 
model behavior \citep{adebayo2018sanity, 
hooker2019benchmark}; our work extends this to 
architectural-level causal measurement, asking whether 
gradient identifies which components drive OOD 
generalization. 
The failure is distinct: gradient is structurally 
inverted at the component level, explained by redundant 
circuit topology rather than input correlations. Prior 
work also identifies heads with degenerate patterns 
\citep{michel2019sixteen, sun2024massive}; our finding 
is distinct in using gradient magnitude, providing 
causal validation, and showing Bloats are individually 
dispensable but collectively critical.

\section{Limitations and Future Work}

Our findings raise three questions for future work. First, does the layer-wise Hero/Bloat structure persist in pretrained LLMs such as GPT-2 or LLaMA? TransformerLens-based replication would directly address whether our taxonomy scales beyond trained-from-scratch models. Second, does the taxonomy replicate under activation patching? This stronger, distribution-preserving counterfactual may reveal additional structure or refine borderline classifications. Third, is early-layer Bloat concentration a signature of shortcut circuit formation? Tracking Hero and Bloat emergence across training checkpoints would test whether the gradient-causal gap reflects the temporal dynamics of grokking directly \citep{nanda2023progress}.

\section{Conclusion}

Gradient magnitude fails to identify causally important 
transformer components on algorithmically complex tasks: 
early-layer Gradient Bloats are individually redundant 
but collectively critical, while late-layer Hidden 
Heroes are causally indispensable despite low gradient 
signal. The layer-wise organization of this failure 
suggests a division of labor during training between 
early-layer feature extraction and late-layer 
algorithmic computation that gradient attribution 
systematically inverts. Circuit analyses and pruning 
methods that rely on gradient attribution may 
systematically overlook the components most responsible 
for generalization, motivating causal validation as a 
prerequisite for architectural claims.

\bibliography{paper}

@article{elhage2021mathematical,
  title={A Mathematical Framework for Transformer Circuits},
  author={Elhage, Nelson and Nanda, Neel and Olsson, Catherine 
          and Henighan, Tom and Joseph, Nicholas and Mann, Ben 
          and Askell, Amanda and Bai, Yuntao and Chen, Anna 
          and Conerly, Tom and DasSarma, Nova and Drain, Dawn 
          and Ganguli, Deep and Hatfield-Dodds, Zac and 
          Hernandez, Danny and Jones, Andy and Kernion, Jackson 
          and Lovitt, Liane and Ndousse, Kamal and Amodei, Dario 
          and Brown, Tom and Clark, Jack and Kaplan, Jared and 
          McCandlish, Sam and Olah, Chris},
  journal={Transformer Circuits Thread},
  year={2021},
  note={https://transformer-circuits.pub/2021/framework/index.html}
}

@inproceedings{nanda2023progress,
  title={Progress measures for grokking via mechanistic interpretability},
  author={Nanda, Neel and Chan, Lawrence and Lieberum, Tom and Smith, Jess and Steinhardt, Jacob},
  booktitle={The Eleventh International Conference on Learning Representations},
  year={2023}
}

@inproceedings{sun2023wanda,
  title={A Simple and Effective Pruning Approach for Large Language Models},
  author={Sun, Mingjie and Liu, Zhuang and Bair, Anna and Kolter, J Zico},
  booktitle={The Twelfth International Conference on Learning Representations},
  year={2024}
}

@inproceedings{adebayo2018sanity,
  title={Sanity Checks for Saliency Maps},
  author={Adebayo, Julius and Gilmer, Justin and Muelly, Michael 
          and Goodfellow, Ian and Hardt, Moritz and Kim, Been},
  booktitle={Advances in Neural Information Processing Systems},
  volume={31},
  year={2018}
}

@article{han2015learning,
  title={Learning both Weights and Connections for Efficient 
         Neural Networks},
  author={Han, Song and Pool, Jeff and Tran, John and Dally, 
          William J},
  journal={Advances in Neural Information Processing Systems},
  year={2015}
}

@inproceedings{michel2019sixteen,
  title={Are Sixteen Heads Really Better than One?},
  author={Michel, Paul and Levy, Omer and Neubig, Graham},
  booktitle={Advances in Neural Information Processing Systems},
  volume={32},
  year={2019}
}

@article{sun2024massive,
  title={Massive Activations in Large Language Models},
  author={Sun, Mingjie and Liu, Xinlei and Bair, Alec 
          and Kolter, J. Zico},
  journal={arXiv preprint arXiv:2402.17762},
  year={2024}
}

@inproceedings{wang2022interpretability,
  title={Interpretability in the Wild: a Circuit for Indirect Object Identification in GPT-2 small},
  author={Wang, Kevin and Variengien, Alexandre and Conmy, Arthur and Shlegeris, Buck and Steinhardt, Jacob},
  booktitle={The Eleventh International Conference on Learning Representations},
  year={2023}
}

@inproceedings{conmy2023automated,
  title={Towards Automated Circuit Discovery for Mechanistic 
         Interpretability},
  author={Conmy, Arthur and Mavor-Parker, Augustine and Lynch, 
          Aengus and Heimersheim, Stefan and Garriga-Alonso, 
          Adri{\`a}},
  booktitle={Advances in Neural Information Processing Systems},
  year={2023}
}

@inproceedings{varma2023explaining,
  title={Explaining Grokking Through Circuit Efficiency},
  author={Varma, Vikrant and Shah, Rohin and Kenton, Zachary 
          and Kram{\'a}r, J{\'a}nos and Kumar, Ramana},
  booktitle={International Conference on Learning Representations},
  year={2024}
}

@inproceedings{hooker2019benchmark,
  title={A benchmark for interpretability methods in deep neural networks},
  author={Hooker, Sara and Erhan, Dumitru and Kindermans, Pieter-Jan and Kim, Been},
  booktitle={Advances in Neural Information Processing Systems},
  volume={32},
  year={2019}
}
\bibliographystyle{icml2026}

\section*{Impact Statement}
This paper presents work whose goal is to advance the 
field of Machine Learning. We identify a systematic 
failure mode in gradient-based attribution methods 
widely used in mechanistic interpretability, with the 
primary consequence being methodological: circuit 
analyses and pruning pipelines that rely on gradient 
attribution may systematically overlook components 
responsible for model generalization. This work was 
conducted independently without external funding. 
There are no additional societal consequences we feel 
must be specifically highlighted here.


\clearpage
\appendix

\section{Hyperparameters and Training Details}
\label{sec:appendix}
\begin{table}[h]
\centering
\caption{Model and training configuration.}
\begin{tabular}{ll}
\toprule
\textbf{Parameter} & \textbf{Value} \\
\midrule
Layers ($L$) & 4 \\
Heads ($H$) & 4 \\
$d_{\text{model}}$ & 128 \\
$d_{\text{ff}}$ & 512 \\
Vocabulary size & 104 \\
Batch size & 64 \\
Learning rate & $10^{-3}$ (Adam) \\
Max training steps & 15{,}000 \\
Target train accuracy & 90\% \\
OOD accuracy window & $[20\%, 75\%]$ \\
Gradient batches ($B$) & 50 \\
Seeds & 42, 123, 456, 789, 1010, \\
      & 2020, 3030, 4040, 5050, 6060 \\
\bottomrule
\end{tabular}
\end{table}

Token vocabulary uses integers $\{1,\dots,99\}$ plus 
four special tokens: \texttt{START}, \texttt{SEP}, 
\texttt{END}, \texttt{PAD}. Training sequences have 
lengths $\in [3,7]$; OOD evaluation uses lengths 
$\in \{8,9,10,11\}$. One reversal seed (4040) failed 
to reach a valid OOD accuracy window and was excluded, 
yielding 9 reversal seeds and 10 sorting seeds.

\section{Full Per-Seed Results}
\label{app:fullseeds}
\begin{table}[h]
\centering
\caption{Per-seed Spearman $\rho$ (mean ablation) 
for both tasks.}
\small
\begin{tabular}{lrr}
\toprule
\textbf{Seed} & \textbf{Reversal $\rho$} & 
\textbf{Sorting $\rho$} \\
\midrule
42   & 0.75 & 0.55 \\
123  & 0.72 & 0.64 \\
456  & 0.81 & 0.00 \\
789  & 0.71 & 0.21 \\
1010 & 0.82 & 0.26 \\
2020 & 0.59 & $-$0.18 \\
3030 & 0.82 & 0.41 \\
4040 & --- (skipped) & 0.15 \\
5050 & 0.64 & 0.49 \\
6060 & 0.62 & 0.20 \\
\midrule
\textbf{Mean} & \textbf{0.72} & \textbf{0.27} \\
\textbf{Std}  & \textbf{0.08} & \textbf{0.24} \\
\bottomrule
\end{tabular}
\end{table}

\label{app:superadditivity}
\begin{table}[b!]
\centering
\caption{Individual and joint Bloat ablation accuracy 
drops on sorting. Each row shows the accuracy drop 
when ablating each Bloat individually versus jointly. 
The $14\times$ superadditivity ratio confirms 
redundant circuit compensation.}
\small
\begin{tabular}{lrrr}
\toprule
\textbf{Seed} & \textbf{Baseline} & 
\textbf{Indiv. Drop} & \textbf{Joint Drop} \\
\midrule
42   & 78.0\% & 9.0\%  & 14.0\% \\
123  & 78.0\% & 4.0\%  & 52.0\% \\
456  & 65.0\% & 9.5\%  & 41.0\% \\
789  & 64.0\% & 1.3\%  & 36.0\% \\
1010 & 66.0\% & 2.5\%  & 61.0\% \\
2020 & 67.0\% & 3.6\%  & 53.0\% \\
3030 & 72.0\% & $-$1.3\% & 10.0\% \\
4040 & 73.0\% & $-$0.5\%& 73.0\% \\
5050 & 72.0\% & 1.0\%  & 26.0\% \\
6060 & 72.0\% & 2.7\%  & 72.0\% \\
\midrule
\textbf{Mean} & & \textbf{3.1\%} & \textbf{43.8\%} \\
\textbf{Ratio} & & & \textbf{14.1$\times$} \\
\bottomrule
\end{tabular}
\end{table}

\clearpage

\begin{figure}[H]
    \includegraphics[width=0.9\textwidth]{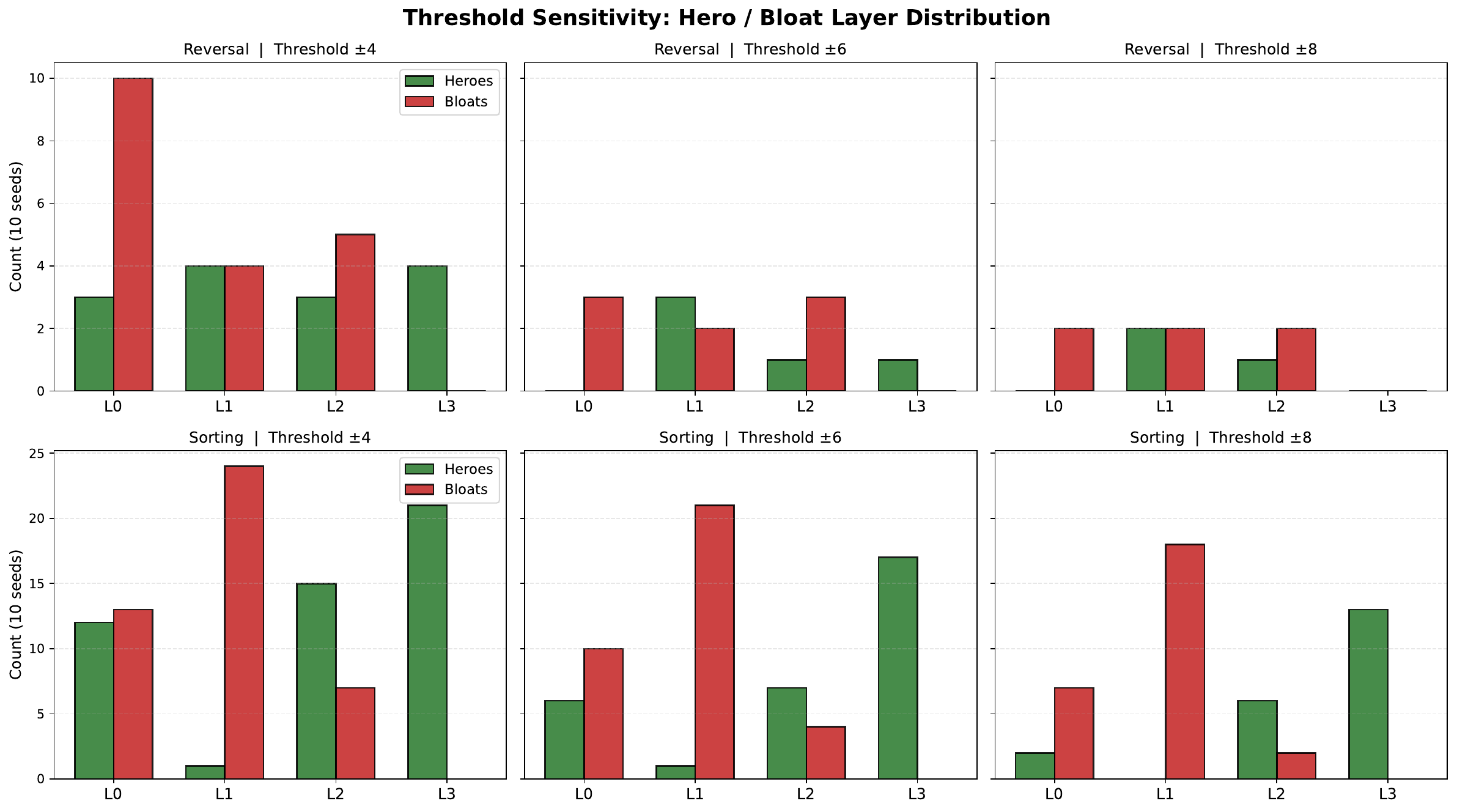}
    \caption{Hero/Bloat layer distribution at 
    thresholds $\pm4$, $\pm6$, $\pm8$. Layer~1 
    consistently accumulates Bloats and Layer~3 
    consistently accumulates Heroes across all 
    thresholds, confirming the pattern is not an 
    artifact of threshold choice.}
    \label{fig:threshold}
\end{figure}
\section{Threshold Sensitivity}
\label{app:threshold}

\begin{table}[h]
\centering
\caption{Hero and Bloat counts per layer at three 
thresholds on the sorting task (summed across 10 
seeds, attention heads only). The layer-wise 
structure --- Layer~1 accumulating Bloats, Layer~3 
accumulating Heroes --- is stable across all 
thresholds.}
\small
\begin{tabular}{lrrrrrr}
\toprule
& \multicolumn{2}{c}{$\pm4$} 
& \multicolumn{2}{c}{$\pm6$} 
& \multicolumn{2}{c}{$\pm8$} \\
\textbf{Layer} & H & B & H & B & H & B \\
\midrule
0 & 12 & 13 & 6  & 10 & 2  & 7  \\
1 & 1  & 24 & 1  & 21 & 0  & 18 \\
2 & 15 & 7  & 7  & 4  & 6  & 2  \\
3 & 21 & 0  & 17 & 0  & 13 & 0  \\
\bottomrule
\end{tabular}
\end{table}

\clearpage

\section{Layer Gradient Norm Distribution}
\label{app:gradnorms}

\begin{figure}[h!]
    \centering
    \includegraphics[width=0.9\textwidth]{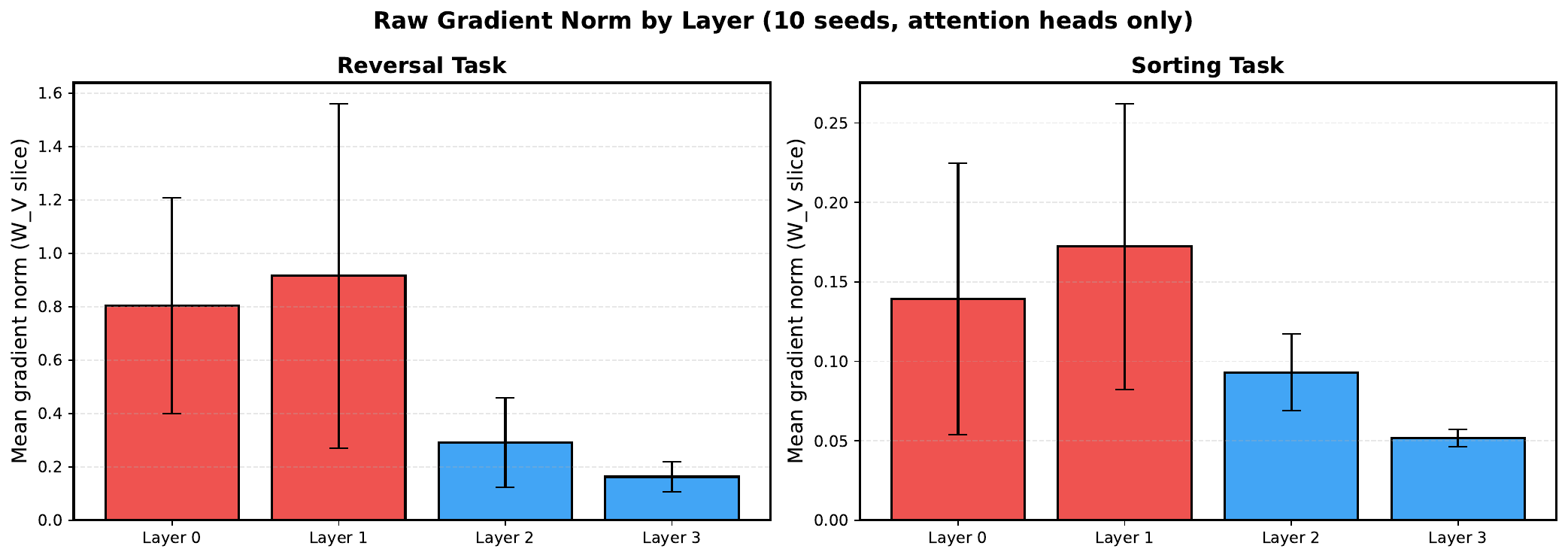}
    \caption{Mean gradient norm ($W_V$ slice) per 
    layer across 10 seeds. On sorting (right), norms 
    decrease monotonically after Layer~1 
    (0.139, 0.172, 0.093, 0.052), confirming that 
    early-layer components accumulate higher gradient 
    signal regardless of causal importance. The 
    reversal task (left) shows the same monotonic 
    decrease but with higher absolute values.}
    \label{fig:gradnorms}
\end{figure}

\section{Attention Pattern Visualization}
\label{app:attention}

\begin{figure}[h!]
    \centering
    \includegraphics[width=0.9\textwidth]{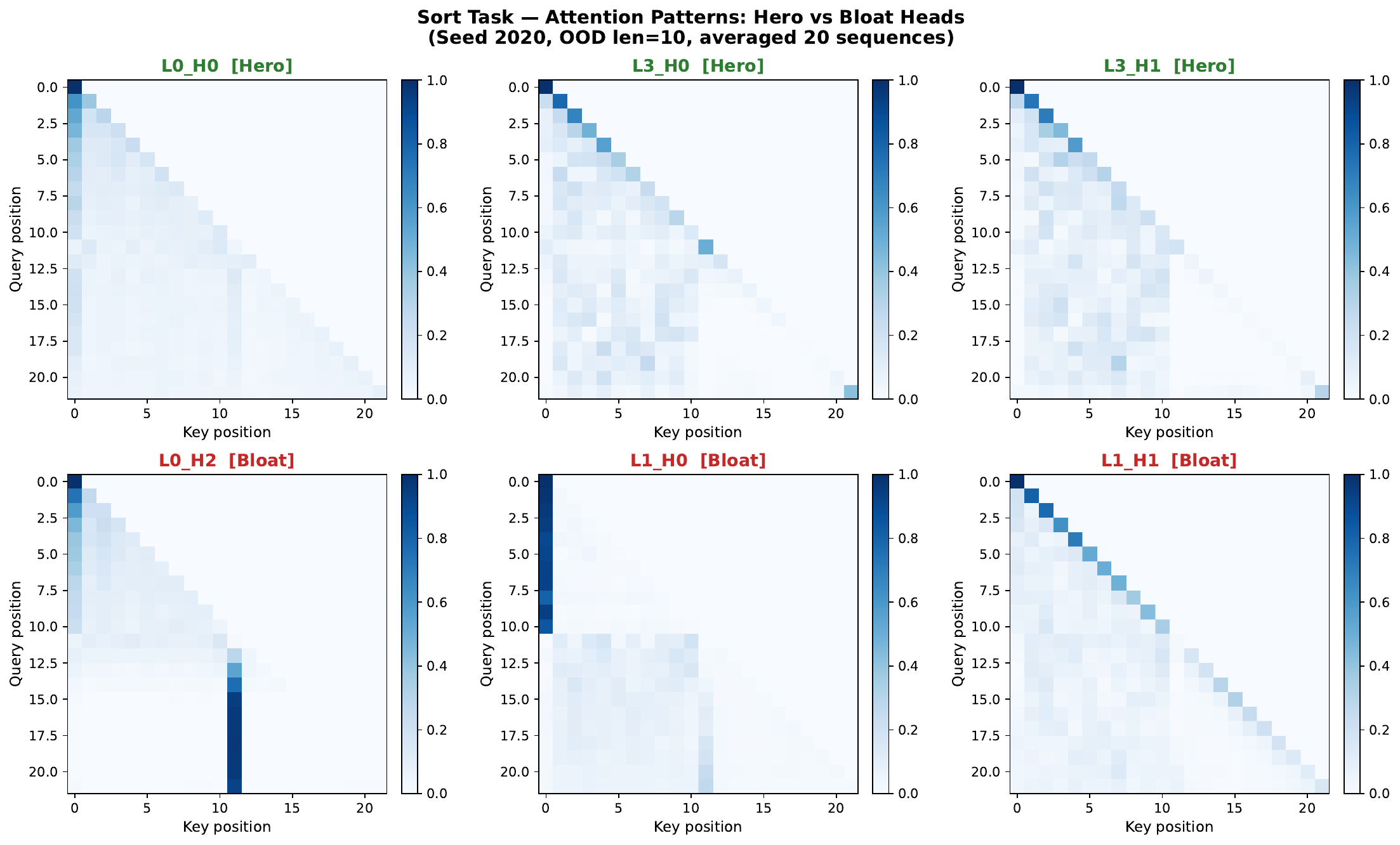}
    \caption{Attention patterns for representative 
    Hero and Bloat heads on the sorting task (Seed 
    2020, OOD length 10, averaged over 20 sequences). 
    Hero heads (L3\_H0, L3\_H1) show structured, 
    position-sensitive attention; Bloat heads 
    (L1\_H0, L1\_H1) show diffuse or sink-token 
    patterns consistent with broad feature extraction.}
    \label{fig:attention}
\end{figure}

\clearpage

\section{All-Seed Scatter Plots}
\label{app:allseeds}

\begin{figure}[h]
    \centering
    \includegraphics[width=\textwidth]{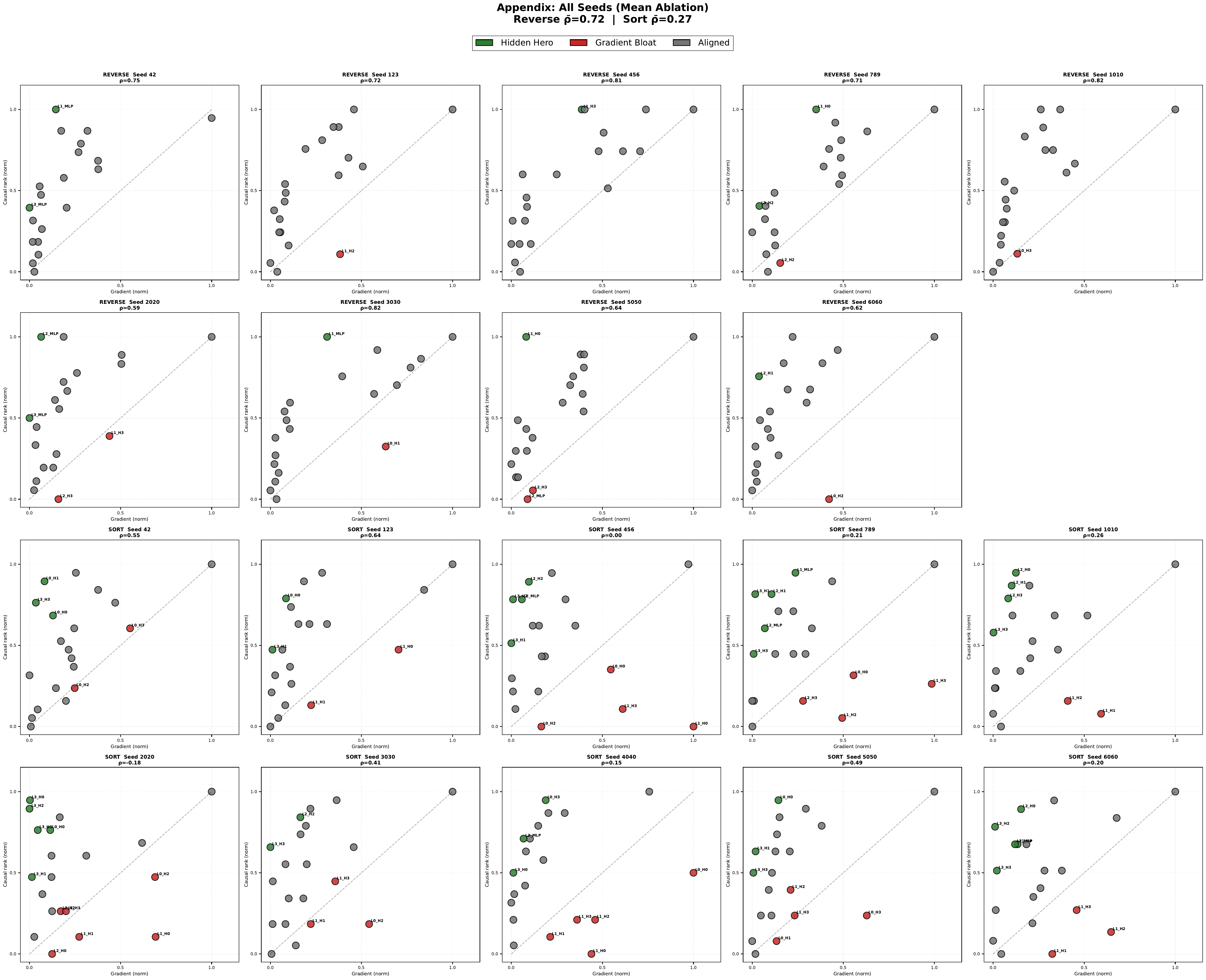}
    \caption{Gradient magnitude (norm) vs.\ causal 
    importance rank for all seeds, both tasks, mean 
    ablation baseline. Green = Hidden Hero; Red = 
    Gradient Bloat; Grey = Aligned. The collapse in 
    correlation on sorting is visible across seeds, 
    with Seed 456 ($\rho=0.00$) and Seed 2020 
    ($\rho=-0.18$) showing the most severe failures.}
    \label{fig:allseeds}
\end{figure}

\end{document}